\documentclass{article}

\usepackage{arxiv}

\usepackage[utf8]{inputenc} 
\usepackage[T1]{fontenc}   
\usepackage{graphicx}
\usepackage{amsmath}
\usepackage{amsfonts}
\usepackage{color,soul}
\usepackage{subfigure}
\usepackage{algorithm}
\usepackage{algorithmic}
\usepackage{authblk}
\usepackage[labelfont=bf]{caption}
\usepackage{hyperref}
\newtheorem{proposal}{Proposal}

\begin{document}
\title{MLDev: Data Science Experiment Automation and Reproducibility Software}

\author[]{\textbf{Anton~Khritankov}$^1$}
\author[]{\textbf{Nikita~Pershin}}
\author[]{\textbf{Nikita~Ukhov}}
\author[]{\textbf{Artem~Ukhov}}
\affil[]{Moscow Institute of Physics and Technology \\ Dolgoprudny, Moscow Region, Russian Federation \\ 
$^1$\texttt{anton.khritankov@phystech.edu}}
\date{}

\hypersetup{
pdftitle={MLDev: Data Science Experiment Automation and Reproducibility Software},
pdfsubject={cs.LG},
pdfauthor={Anton~hritankov, Nikita~Pershin, Nikita~Ukhov,  Artem~Ukhov},
pdfkeywords={experiment automation, data science, reproducibility},
}

\maketitle           
\begin{abstract}
In this paper we explore the challenges of automating experiments in data science. We propose an extensible experiment model as a foundation for integration of different open source tools for running research experiments. We implement our approach in a prototype open source MLDev software package and evaluate it in a series of experiments yielding promising results. Comparison with other state-of-the-art tools signifies novelty of our approach.

\keywords{experiment automation, data science, reproducibility}
\end{abstract}

\section{Introduction}

The ability to reproduce results and use them in future work is one of the key expectations from the modern data science. Herewith, amount of experimental and empirical papers significantly exceeds amount of theoretical publications as shown in the analysis \cite{gundersen2018reproducible,pineau2020improving}. Despite the demand and recent progress there are still unsolved problems that hinder further development, reduce trust level and the quality of results \cite{hutson2018artificial}.

Gundersen et al. \cite{gundersen2018reproducible} study 385 papers containing empirical results from AAAI and IJCAI conferences. More than two third of publications included experiment design, more than a half of the papers included the pseudo code of the algorithm, the problem statement and the training data. At the same time, research questions, purposes of the study, hypotheses tested, source code and detailed analysis of results are often not included in the published papers. Authors indicate that lack of this information significantly influence reproducibility of the research.

Results of the roundtable \cite{roundtable} highlight the reproducibility problem and indicate that a solution of problem requires use of software tools as well as inclusion of topics on experiment design in research training programs in data science.

In 2018 and 2019, organisers of NeurIPS and ICLR conferences \cite{pineau2019iclr,pineau2020improving} explored instruments that can be used to increase reproducibility of research. They offered authors the checklists for self-assessment before submitting articles, suggested to author source code and instructions to reproduce, invited submissions with reproduction of previous researches \cite{pineau2020improving}.

In this paper we describe our approach to improving reproducibility. We suggest to extract the definition of an experiment from the program code and the paper text and define it in both machine and human-readable form. Such specifications of the experiment should be sufficient to reproduce and automate routine tasks. In order to check our idea, we implement a prototype of MLDev system and test it on several examples.

The main contributions of the paper are as follows. First, we derive quality attributes for data science experiment automation and reproducibility software. Then we propose a new approach to automated execution of experiments based on experiment specification and evaluate it against the requirements and other tools. We also open source the implementation of the prototype MLDev system so that other researches could evaluate our approach.

In the next section, we specify the reproducibility problem. In Section \ref{sect:solution} we will describe the proposed approach based on separation of experiment specification in a standalone artifact. In Section \ref{sect:evaluation} we present the results of empirical evaluation of suggested approach and analysis of the obtained results. Section \ref{sect:related-work} includes description of software similar in scope and points out difference with the proposes. 

\section{Problem statement}
\label{sect:statement}

An \emph{experiment} is a procedure carried out to support or refute research hypotheses. Examples of such research hypotheses in data science could be existence of tendencies in data, the choice of model parameters or that one model is not the same as another. 

Experiment design is not as simple as it may look. Even a basic data science experiment with random permutation and data splitting into train and test with many such trials exhibits randomization as a design principle to reduce confounding. A structure of the experiment commonly includes the goal, design choices, a list of hypotheses and their acceptance and rejection criteria, source data and expected results. If an experimental procedure is given as a sequence of stages, it is said that an \emph{experiment pipeline} is defined. After running the pipeline, measurements are analysed and conclusions whether hypotheses can be refuted are drawn.

We use definitions of different types of reproducibility suggested by NISO and ACM \cite{NISO}:
\begin{itemize}
\item \emph{Repeatability} (Same team, same experimental setup). The same measurements can be obtained by the same researcher within the specified error margin using the same procedure and measurement system, under the same conditions. For computational experiments, this means that a researcher can reliably repeat her own computation.

\item \emph{Reproducibility} (Different team, same experimental setup) An independent group can obtain the same result using the author’s own artifacts.

\item \emph{Replicability} (Different team, different experimental setup) An independent group can obtain the same results using artifacts which they develop completely independently.
\end{itemize}

Our goal with the MLDev project is to develop software and supporting methodology to help ensure reproducibility, that is, reusability of experiments and results among researchers. Based on the results of the preliminary literature review, we indicate the identified sources of non-reproducibility \cite{gundersen2018reproducible,pineau2019iclr,pineau2020improving,JupyterRepro}:
\begin{itemize}
  \item View of the source code as an auxiliary result. Industrial software development methods are not used. These factors result in the code defects and distortion of results as stated by Storer et al. \cite{storer2017bridging}.
  \item Insufficient configuration management, inability to reproduce conditions and procedures of the experiment. This includes execution environment, external dependencies, unavailability of data or source code \cite{JupyterRepro}.
  \item Lack of documentation and insufficient description of experiment \cite{trisovic2021large,gundersen2018reproducible}.
\end{itemize}

Many of these reasons are related to the area of software engineering, a discipline which is not a major for data scientists. Others are related to the willingness to publish the results faster and are most likely caused by the violation of empirical research methodology.

We additionally interviewed heads of data analysis laboratories, academics, students and software developers at Moscow Institute of Physics and Technology, Higher School of Economics and Innopolis university in order to elicit requirements for this type of software. As a result, we highlight the following quality characteristics:
\begin{itemize}
  \item \emph{REQ1. Extensibility.} Sufficient functionality and extensibility to define and execute a wide range of experiments in a reproducible manner.
  \item \emph{REQ2. Clarity.} A system should be easy to use and doesn't result in unclear errors or significant increase in time for research.
  \item \emph{REQ3. Compatibility.} Compatibility with existing libraries and tools for conducting computational experiments.
  \item \emph{REQ4. No lock-in.} Freedom from risk of impossibility of publishing the obtained results or difficulties during the process.
\end{itemize}

The complete list of requirements and quality attributes in accordance with ISO/IEC 25010 standard provided in the Appendix \ref{sect:requirements}.

In the next section we propose several architectural decisions, which help overcome the stated reasons of non-reproducibility.

\section{Proposed solution}
\label{sect:solution}

\subsection{Experiment specification}
\label{sect:sxperiment-spec}

Following the Model-Driven Development and Language-Oriented Programming approaches \cite{dmitriev2004language,voelter2018fusing} we address functional extensibility (REQ1) by defining a separate model for the experiment. The specification of the experiment captures the structure of the experiment and serves as a basis for integration of external tools and data with the user code.

\begin{proposal} 
\label{pro:model}
Introduce experiment specification as a separate artifact from the source code and publication. Provide a core conceptual model of the experiment and means for users to extend it in their experiment specifications.
\end{proposal}

The resulting experiment specification allows for the following:
\begin{itemize}
  \item Gathering all information that is crucial for reproducibility in one place according to a common meta-model provided by the software.
  \item Use of automated tools for the analysis of the experiment design analysis and generation of results, which is common for language-oriented programming.
  \item Integration of external tools, both open source and those providing open interfaces.
\end{itemize}

\paragraph{Instance-based composition model.} We propose to define experiment specification as an object-oriented model with instance-based composition instead of class inheritance. In such a model users can add objects with user-defined types and compose new objects from existing ones. The model also supports untyped objects, that is objects use of which does not require specification of a type. 

This approach allows users to reuse state and behaviour by composition, while on the other hand it allows to avoid complexity with polymorphism and virtual inheritance. Indeed, we specifically restrict use of class-based inheritance to tool and plugin providers who are professional software engineers. Indeed, object-oriented modeling and design competencies are not widespread in the research community and it is still not known whether user-defined class based models are well-suited for experiment design. 

\paragraph{Pipeline as a polyforest.} The experiment specification also defines the order of the computation for many hypothesis included in the experiment. Unlike directed computational graphs (pipelines) often used for this purpose, the computational oriented forest (polyforest) seems to fit better. Within an experiment it is necessary to check several hypotheses, algorithms to check each hypothesis are represented as graphs with overlapping vertices. Moreover, an experiment execution context includes the execution order and the usage of services that accompany the execution of experiment. 

The resulting conceptual model is shown at Fig.~\ref{fig:experiment-model} using the concept map notation. Let's describe each concept described on this map.

\emph{Data.} In order to ensure better control over the experiment results we include data versioning control, therefore all the inputs and outputs are versioned on every execution.

\emph{Stages.} An experiment is divided into stages. Inside stages researcher defines the inputs and outputs, what are the execution parameters. Further, the stages are grouped into pipelines, which makes it possible to effectively use the computation forest model.

\emph{Algorithm.} The procedure of the experiment is designed so that it tests the hypotheses. It is determined by the source code and the order of stages execution. The procedure may invoke other algorithms and user source code.

\emph{Hypotheses.} Hypotheses are tested according to a specific procedure and the results are saved together with execution logs and reports. Computational forest model enables setting up several experiment scenarios to test more then one hypothesis.

\emph{Reports and execution log.} Reports and execution log are needed for control of the runs and analyzing the results of the execution.

\emph{Dependencies.} Dependencies include external software and data that needs to be set up before experiment is run. MLDev creates a virtual environment and installs all the dependencies needed according to specified versions. This process allows to have similar execution environment on another computer and easily reproduce experiment.

\emph{Services.} Services provide more abilities to control experiment execution as well as add user-defined services with specific functions. MLDev provides data versioning control, telegram notification bot and Tensorboard logging of the experiment execution. See below for more details.

\subsection{Extensibility mechanisms}
\label{sect:extensibility-mech}

In order to support extensibility (REQ1) provided by the language-oriented programming paradigm and Proposal \ref{pro:model} we apply the Microkernel architectural pattern twice. First, we separate MLDev kernel - the base module that is able to interpret and run experiments specifications, and plugins or extras that enhance the MLDev functionality. Second, we provide a set of templates that provide other artifacts specific to the kind of experiments the user is going to develop and run.

\begin{proposal}
\label{pro:plugins}
Introduce MLDev core and plugins that supply implementation to types and objects defined in the experiment specification. Use templates for experiment repositories to provide specific artifacts.
\end{proposal}

\paragraph{Microkernel architecture.} Microkernel architecture of MLDev also helps in reducing complexity of the system being deployed and run as unnecessary dependencies are not included. Thus we also address REQ2 - Clarity. See Table \ref{tabl:opensource} for a list of plugins and open source technologies they use to implement extensions to MLDev base. Implementation of MLDev system as open-source project with dedicated core and template examples help users to adapt system to their needs.

\paragraph{Templates library.} A library of the predefined templates lowers entry threshold for executing a reproducible experiment and saves the costs spent on preparation and presentation of the results with the help of partial automation and standardisation. Wherein, template can be adapted for the researchers needs.

\begin{figure}
    \centering
    \begin{subfigure}
        \centering
        \begin{minipage}[t]{0.47\textwidth}
            \includegraphics[scale=0.17, trim=40 140 50 190, clip]{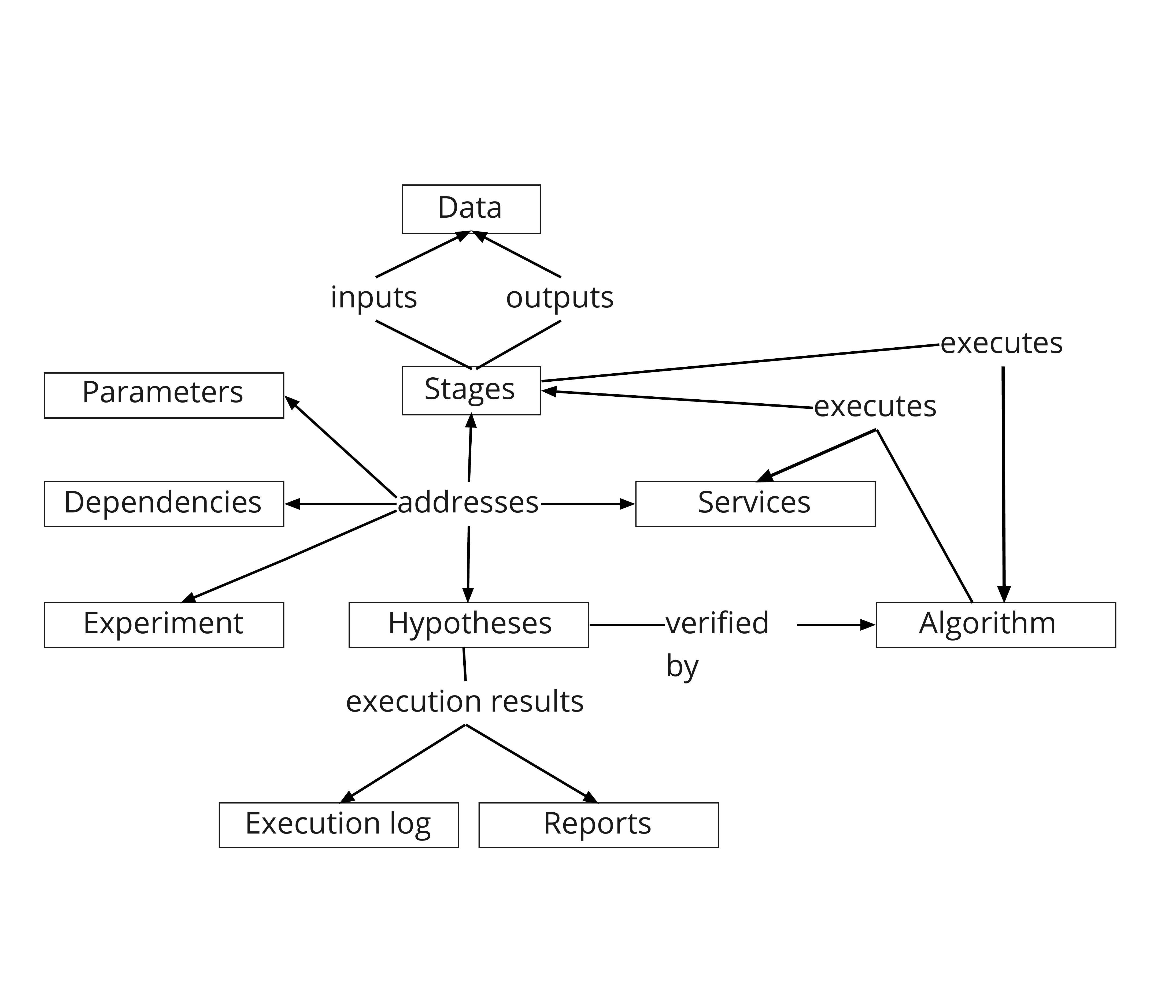}
            \caption{A conceptual model of the experiment and basic types provided by MLDev. More concepts can be added by plugins and templates. Key: Concept map.}
            \label{fig:experiment-model}
        \end{minipage}
    \end{subfigure}
    \begin{subfigure}
        \centering
        \begin{minipage}[t]{0.47\textwidth}
        \includegraphics[scale=0.16, trim=0 50 0 110, clip]{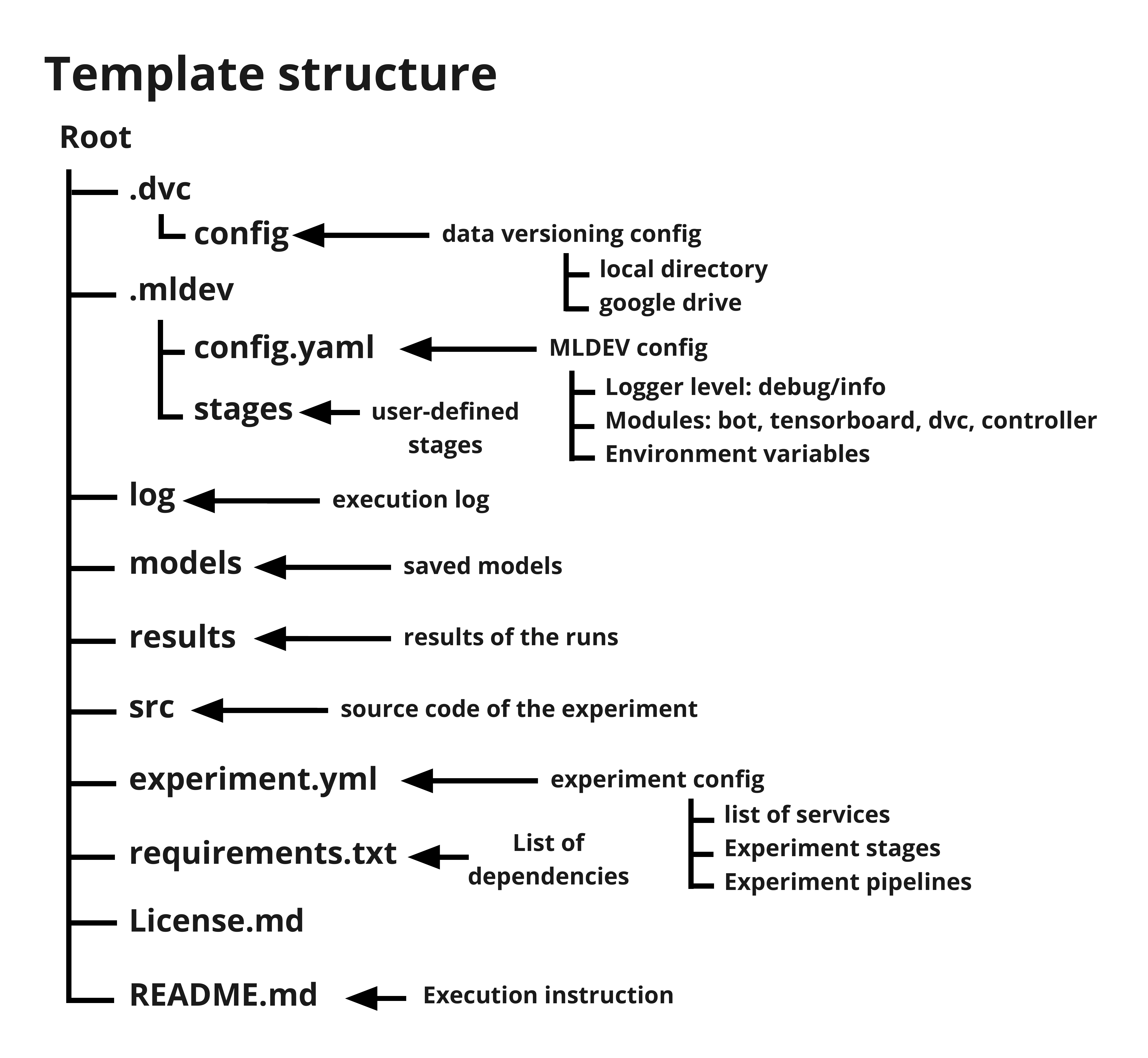}
        \caption{An example of a template with description of structural elements used by MLDev to provide artifacts for reproducible experiments.}
        \label{fig:template-structure}
        \end{minipage}
    \end{subfigure}

\end{figure}

The initial template includes the experiment design, the artifacts and reports mockups, which further will be filled with the results of runs. See Fig. \ref{fig:template-structure} for more details. Template can also provide user-defined data types for experiment specification extending the original object model. Examples of usage will be described further.

\subsection{Open source development}
\label{sect:open-source-dev}

Nowadays, a common approach to implement publicly available tools is running an open source project. The open source development approach, if done right, results in larger participation and cooperation with the target audience. In addition, by running the MLDev project as open source and integrating with other open source tools we address REQ2 by supporting open reviews and continuous community testing, REQ3 through increased number of applications and compatibility tests and REQ4 with an option to fork the project and continue without vendor lock-in. As most open source data science libraries are written in Python, we also implement MLDev in Python thus increasing interoperability (REQ3)

\begin{proposal} 
\label{pro:opensource} 
Open sourcing the project allows for integration of the other open source technologies that increase automation and reproducibility levels, increase overall quality and functionality.
\end{proposal}

\paragraph{Open source integration.} We leverage available open source tools to extend functionality of MLDev. This is an architectural decision to prefer integration through CLI and API instead of reimplementing external tools or reusing at the source code level. This provides for lower development costs, better agility and upgradeability although at the cost of efficiency and flexibility. See Table \ref{tabl:opensource} for a list of open source tools used by MLDev in addition to standard libraries.

\begin{table}[h!]
    \centering
    \caption{MLDev uses open source tools and libraries to extend its functionality via plugins.}
    \medskip
    \label{tabl:opensource}    
    \begin{tabular}{ |p{5cm}|p{3cm}|p{3cm}|  }
    
    \hline
    Feature & Implemented with & Plugin \\
    \hline
    \multicolumn{3}{|c|}{Configuration management and version control} \\
    \hline
    Source code & Git & mldev-dvc \\
    Data and results & DVC \cite{DVC} & mldev-dvc \\
    External libraries & python venv & mldev \\
    \hline
    \multicolumn{3}{|c|}{Experiment logging and debugging} \\
    \hline
    Metrics and measurements & Tensorboard & mldev-tensorboard \\
    Text logs & None & mldev \\
    Notifications & Telegram Bot & mldev-bot \\
    Results demo & Flask & mldev-controller \\
    \hline
    \multicolumn{3}{|c|}{Reproducibility testing} \\
    \hline
    Unit testing & pytest, hypothesis & mldev-test* \\
    Code quality & flake8, black & mldev-test* \\
    Reproducibility testing & gitlab-runner CI & mldev \\ 
    \hline
    \multicolumn{3}{|c|}{Visualization and presentation} \\
    \hline    
    Notebook support & ipython, Jupyter & mldev-ipython* \\ [1ex]
    \hline
    \end{tabular}

\end{table}
\section{Experimental evaluation}
\label{sect:evaluation}

\subsection{Experiment design}
\label{sect:experiment-design}

In order to test our proposals in solving experiment automation and reproducibility problem, we implemented a prototype of the MLDev system  \cite{MLDev}. The goal of the experiments is to test whether our design decisions and MLDev implementation are suitable for real-world experiments. That is, we check feasibility of design so that it could be implemented, and check applicability so that resulting software is suitable for the original experiment automation needs.

In the experiments we check a hypothesis that the MLDev system can be used for creation of the experiments from scratch as well as for reproducing experiments conducted earlier. We measure the following metrics and outcomes:
\begin{itemize}
    \item Was the experiment automation successful?
    \item Time needed to develop experiment specification for MLDev.
    \item How similar are results obtained with MLDev and without it?
    \item Are the results reproducible?
\end{itemize}

For the experiment we selected one paper that uses MLDev to adapt previously implemented experiment to MLDev for the purpose of reproducibility and another paper by different authors to check whether a third-party experiment can also be adapted to run with the same results via MLDev.

Experiments were conducted during Jan-May 2021 using MLDev versions 0.2.dev3 and 0.3.dev1 respectively on Ubuntu 18 (x64) on a laptop and PC and Google Colab.

\subsection{Application to a new experiment}
\label{sect:new-experiment}

The article \cite{khritankov2021hidden} examines the problem of analyzing the quality of continuous machine learning systems, that is systems that use machine learning to implement its major features and that get their predictive models updated over time. The paper presents a simulation experiment for the housing prices prediction problem, in which a feedback loop effect occur when previously made predictions fall into new training data. The original experiment source code was available as a Jupyter Notebook.

In the process of reproducing the results, the following tasks were solved using MLDev v0.2.dev3:
\begin{itemize}
    \item Prepared a public repository for an experiment based on the basic experiment template-default.
    \item The experiment code is extracted from the notebook and is placed according to the repository structure.
    \item Implemented a driver program for running an experiment from the command line.
    \item A description of the MLDev experiment was prepared and the external dependencies of the experiment code were determined for porting to other runtime environments.
    \item Implemented a script for executing an experiment with tunable  hyperparameters to analyze the error margin of the obtained results.
\end{itemize}

While porting the original source code to the MLDev system, the following issues were identified and resolved in the original experiment:
\begin{itemize}
    \item Incomplete initialization of the random number generator (affects repeatability).
    \item Error in the data update algorithm (does not affect the conclusions of the publication).
\end{itemize}

The total effort for creating the repository and transferring experiment to the MLDev system were 4.5 hours, and another 4 hours were required later to refine the experiment, eliminate defects and finalize the visualization of the results. The updated experiment is included in an Arxiv paper \cite{khritankov2021analysis}.

\subsection{Reproducibility study for a published paper}
\label{sect:repro-experiment}

Deepak et al.\cite{nathani-etal-2019-learning} study the problem of predicting relations in knowledge graphs. Authors develop an encoder-decoder architecture that uses a graph neural network as encoder and a convolutional neural network as decoder. The source code is written in Python and run with three CLI commands with parameters.

When transferring the experiment to MLDev, the following problems were identified:
\begin{itemize}
    \item An error was detected when re-creating checkpoints used to save models.
    \item The command to run the experiment was too complex and easy to make mistakes, especially when typing them manually.
    \item There were no configurations for all of the experiment scenarios.
\end{itemize}

In order to repeat the results and transfer the experiment to MLDev v0.3.dev1, the following steps were taken:
\begin{itemize}
    \item Eight experiment pipelines have been implemented. Four of them replicating the original experiment configurations and other four configurations that allow to conduct the experiment with Google Colab \cite{bisong2019google}.
    \item Resolved a problem with creating checkpoints for saving models.
    \item A notebook has been created to repeat the experiment by running just two cells.
    \item A public repository has been prepared with a description of the problem, the result of the experiment and detailed instructions for reproducing the experiment.
\end{itemize}

The total development effort is five hours and about forty hours were spent on training and testing models. The result of the article was reproduced on the main test dataset FB15k \cite{bordes2013translating}. As a result, MLDev solved the problem with the lack of experimental configurations and made it possible to conveniently reproduce each of the pipelines.

\subsection{Analysis and results}
\label{sect:analysis-results}

In both cases the experiments were automated and provided the same results which were published by their authors. Also in both cases results can be easily reproducible through the usage of a single MLDev command.

We were able to confirm that MLDev software is able to run data science experiments with the reproducible results in both cases.

In addition, transferring of the experiments to MLDev helped reveal defects in the source code that might affect the scientific result.

We also confirmed that the experiment specification is extensible enough to accommodate different kinds of experiments: simulation experiments with feedback loops and deep learning experiments.

\section{Related work and comparison}
\label{sect:related-work}

In this section, we compare MLDev with four related approaches. They are selected in accordance with their popularity in research community and whether they support reproducibility. The manual approach is the most basic that researchers use while preparing the experiment for  publication, Jupyter Notebooks are one of the most popular instruments used for ad-hoc research and demonstration of data science experiments \cite{JupyterRepro}. MLFlow and Nextflow are chosen because these projects have existed for a long time, used by researchers in their publications, and these tools are the most representative alternative we could find. Interested reader is also referred to a paper by Isdahl et al.  \cite{isdahl2019out} for a review of different experiment automation tools based on the authors reproducibility evaluation framework.

\subsection{Command line tools}

First, we consider the currently most widespread approach to preparing an experiment for publication, in which all documents and source code necessary for reproducibility of the results are prepared manually. Ensuring reproducibility experiment is associated with many factors and problems described earlier in Section \ref{sect:requirements}. These problems include control of the execution environment, determining the order of the experiment execution, etc. The main problem in this case is that even if the researcher manages to provide all the necessary artifacts and describe the procedure for reproducing the experiment, all these components will not be interconnected, as a result the execution of the experiment will not be easy for the researcher who wants to reproduce it. She will need to execute several commands to prepare the environment and install the libraries, then run the scripts for the experiment, then the scripts to evaluate the result. A good example of this approach to experiment design is a source code repository that accompanies the paper by Bunel et al. \cite{GANRL}, which we were not able to reproduce without modifying the source code.

In general, this approach is susceptible to most of the problems reported by Pineau et al. \cite{pineau2019iclr}, such as lack of data for training, incorrect specification of the training process, errors in the code, and so on. If we scale up an experiment to include multiple executions to test multiple hypotheses, the researchers effort increases superlinearly due to interdependencies between pipelines. Similarly, the confidence in correctness and reproducibility of pipelines diminishes.

\subsection{Jupyter Notebooks}

One of the goals of creating Jupyter notebooks \cite{kluyver2016jupyter} was to ensure reproducibility, but the study by Jiawei W. et al. \cite{JupyterRepro} showed that the presentation of an experiment as a notebook and cells with code brings additional problems related to reproducibility. Nearly 40\% of notebooks rely on functions that employ a random number generator, and their execution showed different results from those published by the authors. Less frequent, but also important problems are errors associated with: 
\begin{itemize}
    \item Using time and date functions.
    \item Using data visualization libraries.
    \item Lack of input data.
    \item Lack of dependency control and configuration management.
\end{itemize}

The main problem associated with the usage of notebooks is the lack of any specified cell execution order. Indeed, notebooks have an option to execute all the cells, but even if the notebook is executed without errors, this will not guarantee the correctness of the results. Thus, designing the experiment as a Jupiter notebook without a clear template and constantly checking the results of the experiment, it is impossible to provide a sufficient level of reproducibility, which is confirmed in a study by Jiawei et al. \cite{JupyterRepro}, where they showed that less than 5\% of notebooks published in open repositories on GitHub provided the expected results.
 
Next, we will look at the tools that can be used to build pipelines and aim to provide reproducibility.

\subsection{NextFlow}

A tool that was developed to run bioinformatics pipelines. Nextflow developers identified the key issues that hinder reproducibility as numerical instability and changes in the runtime when the experiment is ported \cite{di2017nextflow}. NextFlow uses a domain specific language (DSL) and an elaborate meta-model behind the language to define pipelines. A researcher can specify several entry points for the pipeline, thus different scenarios of the experiment can be configured. The problem of numerical instability and changes in the runtime environment is solved by containerizing the stages of the experiment. Therefore, NextFlow can be viewed as a tool to build automated pipelines, while different entry points cannot be considered as specified and easy to use concept for testing several hypotheses.

\subsection{MLflow}

MLflow is a tool designed for commercial data analysis but can also used by researchers. MLflow developers had three main goals when they were developing their tool \cite{zaharia2018accelerating}: 
\begin{itemize}
    \item Experiment execution tracking to provide for provenance.
    \item Configuration definition and reuse.
    \item Model packaging and deployment.
\end{itemize} 

MLflow manages experiments through the concept of ML project. The pipeline is presented as a sequence of stages that are run in a prescribed order. Therefore, it also makes it impossible to test multiple hypotheses within a single experiment configuration. MLflow has many features, but they are more focused on commercial use in production development. This tool can track the execution of the experiment, but due to the fact that it was not developed for research purposes, the inability to record several configurations of an experiment into a single project can become critical when choosing a tool for experiment automation.

\subsection{Analysis and comparison}

We considered four common approaches to implement research experiments ranging from an approach without the use of automation tools to the most elaborate solutions that allow you to control the execution of experiments. In accordance with the requirements described in the section where we define the reproducibility problem, we can conclude that tools such as MLflow and NextFlow provide functionality for solving many problems, which researchers face during the experiment preparation for publication. The first two approaches do not provide appropriate automation means out-of-the-box. Regarding the ease of use, the first two approaches may cause a lot of errors and complicate the control of the experiment \cite{pineau2019iclr,JupyterRepro}. On the other hand detailed data on the use of MLflow and Nextflow is lacking.

MLflow and NextFlow are simple enough to use for a researcher to conduct an experiment by reading the documentation. Integration with already used tools is partially implemented in MLflow and NextFlow, the ability to write your own additional modules was found only in MLflow and requires special knowledge of building python packages. Regarding experiment configuration management, the greatest functionality is provided by MLflow using MLflow tracking and the ability to specify various execution scenarios in the configuration file. The worst approach to configuration management is, of course, the one without using experiment automation systems at all.

When compared to prototype MLDev implementation, the latter tools are more mature and already received attention from the community, Still, their focus on running data analysis and model deployment (MLflow) or running elaborate bioinformatics pipelines is substantially different from the goals of the MLDev project. Incorporating similar ideas and providing an alternative extensible experiment object model MLDev could be more suitable for a wide range of data science research cases. We summarize the comparison results in Table \ref{tab:comparison}.

\begin{table}[]
    \caption{Comparison of MLDev with other approaches with regards to tool requirements, Section~\ref{sect:requirements}}
    \label{tab:comparison}
    \medskip
    \centering
    \begin{tabular}{ |p{2.5cm}|p{3cm}|p{3cm}|p{3cm}|  }
     \hline
    Approach & Experiment model & Extensions & Project type \cite{FOSStype}\\
     \hline
     Manual   & No    & CLI/POSIX tools & -\\
     Jupyter notebook &  No  & Plugins &  Monarchist\\
     MLFlow    & Yes (Fixed) & Plugins, Templates & Corporate\\ 
     Nextflow & Yes (Fixed, DSL) & Plugins & Corporate\\ [0.5ex]
     \hline 
     MLDev    & Yes (Extensible) & Plugins, Model, Templates &  Community\\
     \hline
    \end{tabular}    
\end{table}

Thus, we can conclude that at the moment there are tools that partially solve the problem of reproducibility, but there are also aspects in the process of ensuring reproducibility that are not taken into account by current tools, for example, data versioning, or can be significantly improved: experiment configuration for several execution scenarios, custom modules integration.

\section{Conclusion}

In this paper we study how experiment automation and reproducibility needs of the data science research community could be addressed. Based on literature survey and in-depth interview of the data science professional we propose a novel approach to modeling data science experiments and achieving reproducibility of research.

In general, the creation of the MLdev instrument was motivated by shortcomings of the tools available to researchers and the extensive list of unsolved problems that researchers face when it comes to ensuring reproducibility of their experiments\cite{hutson2018artificial,pineau2019iclr,pineau2020improving}.

We implement a prototype MLDev system and apply it to two experiments thus demonstrating feasibility of design decisions and applicability of approach. Comparison with other tools highlights differences in golas of the projects and their target audiences while distinguishing the MLDev approach to extensible experiment model.


\begin{thebibliography}{10}

\bibitem{gundersen2018reproducible}
Odd~Erik Gundersen, Yolanda Gil, and David~W Aha.
\newblock On reproducible ai: Towards reproducible research, open science, and
  digital scholarship in ai publications.
\newblock {\em AI magazine}, 39(3):56--68, 2018.

\bibitem{pineau2020improving}
Joelle Pineau, Philippe Vincent-Lamarre, Koustuv Sinha, Vincent Larivi{\`e}re,
  Alina Beygelzimer, Florence d'Alch{\'e} Buc, Emily Fox, and Hugo Larochelle.
\newblock Improving reproducibility in machine learning research (a report from
  the neurips 2019 reproducibility program).
\newblock {\em arXiv preprint arXiv:2003.12206}, 2020.

\bibitem{hutson2018artificial}
Matthew Hutson.
\newblock Artificial intelligence faces reproducibility crisis, 2018.

\bibitem{roundtable}
Konstantin Vorontsov, Vladimir Iglovikov, Vadim Strijov, Andrey Ustuzhanin, and
  Anton Khritankov.
\newblock Roundtable: Challenges in repeatable experiments and reproducible
  research in data science.
\newblock {\em Proceedings of MIPT (Trudy MFTI)}, 13(2):90--99, 2021.

\bibitem{pineau2019iclr}
Joelle Pineau, Koustuv Sinha, Genevieve Fried, Rosemary~Nan Ke, and Hugo
  Larochelle.
\newblock Iclr reproducibility challenge 2019.
\newblock {\em ReScience C}, 5(2):5, 2019.

\bibitem{NISO}
{Neil P} {Chue Hong}.
\newblock {\em Reproducibility Badging and Definitions: A Recommended Practice
  of the National Information Standards Organization}.
\newblock National Information Standards Organization (NISO), January 2021.

\bibitem{JupyterRepro}
Jiawei Wang, KUO Tzu-Yang, Li~Li, and Andreas Zeller.
\newblock Assessing and restoring reproducibility of jupyter notebooks.
\newblock pages 138--149, 2020.

\bibitem{storer2017bridging}
Tim Storer.
\newblock Bridging the chasm: A survey of software engineering practice in
  scientific programming.
\newblock {\em ACM Computing Surveys (CSUR)}, 50(4):1--32, 2017.

\bibitem{trisovic2021large}
Ana Trisovic, Matthew~K Lau, Thomas Pasquier, and Merc{\`e} Crosas.
\newblock A large-scale study on research code quality and execution.
\newblock {\em arXiv preprint arXiv:2103.12793}, 2021.

\bibitem{dmitriev2004language}
Sergey Dmitriev.
\newblock Language oriented programming: The next programming paradigm.
\newblock {\em JetBrains onboard}, 1(2):1--13, 2004.

\bibitem{voelter2018fusing}
Markus Voelter.
\newblock Fusing modeling and programming into language-oriented programming.
\newblock In {\em International Symposium on Leveraging Applications of Formal
  Methods}, pages 309--339. Springer, 2018.

\bibitem{DVC}
Data version control tool (dvc).
\newblock \url{https://dvc.org}.
\newblock Accessed: 2021-06-14.

\bibitem{MLDev}
Mldev. an open source data science experimentation and reproducibility
  software.
\newblock \url{https://mlrep.gitlab.io/mldev}.
\newblock Accessed: 2021-06-14.

\bibitem{khritankov2021hidden}
Anton Khritankov.
\newblock Hidden feedback loops in machine learning systems: A simulation model
  and preliminary results.
\newblock In {\em International Conference on Software Quality}, pages 54--65.
  Springer, 2021.

\bibitem{khritankov2021analysis}
Anton Khritankov.
\newblock Analysis of hidden feedback loops in continuous machine learning
  systems.
\newblock {\em arXiv preprint arXiv:2101.05673}, 2021.

\bibitem{nathani-etal-2019-learning}
Deepak Nathani, Jatin Chauhan, Charu Sharma, and Manohar Kaul.
\newblock Learning attention-based embeddings for relation prediction in
  knowledge graphs.
\newblock In {\em Proceedings of the 57th Annual Meeting of the Association for
  Computational Linguistics}, pages 4710--4723, Florence, Italy, July 2019.
  Association for Computational Linguistics.

\bibitem{bisong2019google}
Ekaba Bisong.
\newblock Google colaboratory.
\newblock In {\em Building Machine Learning and Deep Learning Models on Google
  Cloud Platform}, pages 59--64. Springer, 2019.

\bibitem{bordes2013translating}
Antoine Bordes, Nicolas Usunier, Alberto Garcia-Duran, Jason Weston, and Oksana
  Yakhnenko.
\newblock Translating embeddings for modeling multi-relational data.
\newblock In {\em Neural Information Processing Systems (NIPS)}, pages 1--9,
  2013.

\bibitem{isdahl2019out}
Richard Isdahl and Odd~Erik Gundersen.
\newblock Out-of-the-box reproducibility: A survey of machine learning
  platforms.
\newblock In {\em 2019 15th international conference on eScience (eScience)},
  pages 86--95. IEEE, 2019.

\bibitem{GANRL}
Rudy Bunel, Matthew Hausknecht, Jacob Devlin, Rishabh Singh, and Pushmeet
  Kohli.
\newblock Leveraging grammar and reinforcement learning for neural program
  synthesis.
\newblock In {\em International Conference on Learning Representations}, 2018.

\bibitem{kluyver2016jupyter}
Thomas Kluyver, Benjamin Ragan-Kelley, Fernando P{\'e}rez, Brian~E Granger,
  Matthias Bussonnier, Jonathan Frederic, Kyle Kelley, Jessica~B Hamrick, Jason
  Grout, Sylvain Corlay, et~al.
\newblock {\em Jupyter Notebooks-a publishing format for reproducible
  computational workflows.}, volume 2016.
\newblock 2016.

\bibitem{di2017nextflow}
Paolo Di~Tommaso, Maria Chatzou, Evan~W Floden, Pablo~Prieto Barja, Emilio
  Palumbo, and Cedric Notredame.
\newblock Nextflow enables reproducible computational workflows.
\newblock {\em Nature biotechnology}, 35(4):316--319, 2017.

\bibitem{zaharia2018accelerating}
Matei Zaharia, Andrew Chen, Aaron Davidson, Ali Ghodsi, Sue~Ann Hong, Andy
  Konwinski, Siddharth Murching, Tomas Nykodym, Paul Ogilvie, Mani Parkhe,
  et~al.
\newblock Accelerating the machine learning lifecycle with mlflow.
\newblock {\em IEEE Data Eng. Bull.}, 41(4):39--45, 2018.

\bibitem{FOSStype}
John Berkus.
\newblock The 5 types of open source projects.
\newblock
  \url{https://wackowiki.org/doc/Org/Articles/5TypesOpenSourceProjects}.
\newblock Accessed: 2021-06-14.

\end{thebibliography}

\appendix
\section{Quality requirements for experiment automation software}
\label{sect:requirements}

This is a preliminary list of quality requirements for experiment automation and reproducibility software. The requirements are based on series of in-depth interviews of data science researchers, heads of data science laboratories, academics, students and software developers in MIPT and Innopolis university.

Quality categories are given in accordaance with ISO/IEC 25010 quality model standard.

\paragraph{Functionality}
\begin{itemize}
    \item Ability to describe pipelines and configuration of ML experiments.
    \item Run and reproduce experiments on demand and as part of a larger pipeline.
    \item Prepare reports on the experiments including figures and papers.
\end{itemize}

\paragraph{Usability}
\begin{itemize}
    \item Low entry barrier for data scientists who are Linux users.
    \item Ability to learn gradually, easy to run first experiment
    \item Technical and programming skill needed to use experiment automation tools should be lower than running experiments without it.
    \item Users should be able to quickly determine the source of the errors.
\end{itemize}

\paragraph{Portability and compatibility}
\begin{itemize}
    \item Support common ML platforms (incl. Cloud Google Colab), OSes (Ubuntu 16, 18, 20, MacOS) and ML libraries (sklearn, pandas, pytorch, tensorflow…)
    \item Support experiments in Python, Matlab
    \item Run third-party ML tools with command-line interface
\end{itemize}

\paragraph{Maintainability}
\begin{itemize}
    \item Open project, that is everyone should be able to participate and contribute.
    \item Contributing to the project should not require understanding all the internal workings.
    \item Should provide backward compatibility for experiment definitions.
\end{itemize}

\paragraph{Security / Reliability}
\begin{itemize}
    \item Confidentiality of experiment data unless requested by user otherwise (e.g. publish results).
    \item Keep experiment data secure/safe for a long time
\end{itemize}

\paragraph{Efficiency}
\begin{itemize}
    \item Overhead is negligible for small and large experiment compared with the user code.
\end{itemize}

\paragraph{Satisfaction and ease of use}
\begin{itemize}
    \item Must be at least as rewarding / satisfactory / easy-to-use as Jupyter Notebook.
    \item Interface should be similar to other tools familiar to data scientists.
\end{itemize}

\paragraph{Freedom from risk}
\begin{itemize}
    \item Using experiment automation software should not risk having their projects completed and results published.
\end{itemize}

\end{document}